\documentclass[conference]{IEEEtran}
\IEEEoverridecommandlockouts

\usepackage{cite}
\usepackage{amsmath,amssymb,amsfonts}
\usepackage[english]{babel}
\usepackage{graphicx}
\usepackage{textcomp}
\usepackage{xcolor}
\usepackage[hyphens]{url}
\usepackage{hyperref}
\usepackage[normalem]{ulem}
\usepackage{comment}
\usepackage{amsmath}
\usepackage{amssymb}
\usepackage{algorithm,algorithmic}
\def\BibTeX{{\rm B\kern-.05em{\sc i\kern-.025em b}\kern-.08em
    T\kern-.1667em\lower.7ex\hbox{E}\kern-.125emX}}

\begin{document}
%
\title{\Large{\textbf{NMPC-LBF: Nonlinear MPC with Learned Barrier Function for Decentralized Safe Navigation of Multiple Robots in Unknown Environments}}\\
}
%
\author{Amir~Salimi~Lafmejani, Spring~Berman, and Georgios~Fainekos
\thanks{This work was partially supported by DARPA AMP N6600120C4020, NSF CNS 1932068, NSF IIP-1361926, and the NSF I/UCRC CES.}
\thanks{A. Salimi Lafmejani is with the School of Electrical, Computer and Energy Engineering, Arizona State University (ASU), Tempe, AZ, 85287 ({\tt\small asalimil@asu.edu}). 
S. Berman is with the School for Engineering of Matter, Transport and Energy, ASU, Tempe, AZ 85287 ({\tt\small spring.berman@asu.edu}).
G. Fainekos is with the Toyota Research Institute of North America, Ann Arbor, MI 48105 
({\tt\small georgios.fainekos@toyota.com}). 
The work was initiated when G. Fainekos was with the School of Computing and Augmented Intelligence, ASU, Tempe, AZ, 85281. 
}
}
\maketitle
%

\begin{abstract}
In this paper, we present a decentralized control approach based on a Nonlinear Model Predictive Control (NMPC) method that employs barrier certificates for safe navigation of multiple nonholonomic wheeled mobile robots in unknown environments with static and/or dynamic obstacles. This method incorporates a Learned Barrier Function (LBF) into the NMPC design in order to guarantee safe robot navigation, i.e., prevent robot collisions with other robots and the obstacles. We refer to our proposed control approach as NMPC-LBF. Since each robot does not have a priori knowledge about the obstacles and other robots, we use a Deep Neural Network (DeepNN) running in real-time on each robot to learn the Barrier Function (BF) only from the robot's LiDAR and odometry measurements. The DeepNN is trained to learn the BF that separates safe and unsafe regions. We implemented our proposed method on simulated and actual Turtlebot3 Burger robot(s) in different scenarios. The implementation results show the effectiveness of the NMPC-LBF method at ensuring safe navigation of the robots.
\end{abstract}

\section{Introduction}

Collision-free navigation of multiple mobile robots has been extensively studied in the literature for different applications such as autonomous cars~\cite{hou2020development}, warehouse automation~\cite{fan2020distributed}, planetary exploration~\cite{otsu2017look}, and service robots~\cite{majd2021safe}. The existing control approaches for collision-free navigation of multi-robot systems can be categorized as \textit{centralized},~\textit{distributed}, and~\textit{decentralized} methods. Although collision and deadlock avoidance of robots can be guaranteed in centralized control approaches, they suffer from the scalability issue since the computational complexity increases with the number of the robots~\cite{lafmejani2021nonlinear, wang2017safety}. 

In distributed control approaches, inter-robot collision avoidance can be established using inter-robot communication~\cite{wang2017safety,mehrez2017occupancy}. 
%
However, inter-robot communication might be unreliable, or even not possible. 
Thus, decentralized control approaches have been developed to eliminate the aforementioned limitations of centralized and distributed approaches. However, existing decentralized methods are not fully decentralized due to one or more of the following simplifying assumptions: \textbf{(1)} the reliance of robots on inter-robot communication; \textbf{(2)} a priori knowledge of the robots about the positions of obstacles in the environment; \textbf{(3)} dependency of the approach on a global localization system such as a camera or motion capture system; and \textbf{(4)} absence of obstacles in multi-robot scenarios. 

In traditional control methods for collision-free navigation of mobile robots, collision avoidance has been achieved by incorporating the gradient of an artificial potential field into the controller design, e.g., navigation function-based methods~\cite{paternain2017navigation} or attraction-repulsion methods~\cite{khatib1985real}. Recently, several methods have been developed to incorporate collision avoidance into the constraints of an optimization problem in order to enforce the kinematics or dynamics of the robot, constraints on the robot's states and actuation, and the dynamics of the environment~\cite{lafmejani2021nonlinear,wang2017safety,zeng2020safety}. 
For multi-robot systems, collision avoidance is a more challenging task since we need to ensure avoidance of both inter-robot and robot-obstacle collisions. Collision avoidance can be guaranteed by introducing distance functions between each pair of the robots in a centralized framework, as described in~\cite{lafmejani2021nonlinear}. However, using distance functions in a decentralized framework fails to ensure collision avoidance in dynamic environments, since the constraint does not capture the dynamics of unsafe regions. To address this issue, one can use Barrier Functions (BFs) to ensure collision avoidance. These functions separate safe and unsafe regions in the environment and can be included in a constrained minimization problem whose solution minimally deviates from a nominal controller with guaranteed stability~\cite{ames2014control}. 

As a promising control approach for collision-free navigation of mobile robots~\cite{lafmejani2021nonlinear}, Model Predictive Control (MPC) is a powerful feedback control method that computes optimal control solutions by solving a constrained optimization problem over a prediction horizon~\cite{borrelli2017predictive,mehrez2020model}. There are several studies that incorporate BFs into constraints or directly into the objective function of the MPC optimization to ensure collision-free navigation of mobile robots. For instance, a safety-critical MPC method with discrete-time BF has been presented in~\cite{zeng2020safety} for collision-free navigation of mobile robots in known, static environments. The existing BF-based MPC methods analytically construct the BFs and incorporate them into MPC design to ensure collision-free navigation, which is impractical in real-world applications. 
Synthesizing BFs is straightforward in a centralized control architecture where the robots have information about the positions and geometry of the obstacles. Given this information, a BF, representing the boundary of the smallest circle that encloses an obstacle, could be defined for each obstacle. 
On the other hand, it would be challenging for each robot to individually construct BFs in a decentralized manner using only its own sensor measurements {due to the partial observability of the environment and sparsity of the measured data}. 
To address this challenge, several recent studies have investigated the use of neural networks (NNs) to learn BFs offline or online in both known and unknown environments \cite{yaghoubi2020training,long2021learning,zhao2020synthesizing,saveriano2019learning,srinivasan2020synthesis,li2021instantaneous}. 

This paper presents a decentralized NMPC method with learned BF for collision-free navigation of multiple nonholonomic mobile robots in unknown environments.
A learned BF in the form of a Deep Neural Network (DeepNN) is necessary since the map of the environment is unknown and the BF constraints must be incorporated into the NMPC optimization problem.
That is, we need a map from a state of the robot (as used in the NMPC loss function) to the predicted BF value.

The following features of the proposed NMPC-LBF method differentiate it from existing methods for multi-robot navigation: \textbf{(1)} Robots do not require a priori information about the locations or geometry of other robots or obstacles. \textbf{(2)} There is no inter-robot communication that can be used to avoid collisions between robots. \textbf{(3)} Safe navigation is achieved in the presence of non-aggressive dynamic obstacles. \textbf{(4)} Due to its  decentralization, the proposed method can be duplicated on any number of robots for safe navigation. 
\textbf{(5)} 
A global localization system is not required since the method can use on-board sensors of the robot. 
The main novelty of our work compared to approaches for learning BFs in~\cite{yaghoubi2020training,long2021learning,zhao2020synthesizing,saveriano2019learning,srinivasan2020synthesis,li2021instantaneous} is to combine the NMPC
with the learned BF. 
Furthermore, through experiments on actual Turtlebot3 Burger robot(s), we demonstrate that our method is feasible in practice (see~\cite{VideosSimExp}).

\section{Control Problem Formulations}\label{Section:Problem_Formulation}

\subsection{Nonlinear Model Predictive Control (NMPC)}

Our proposed control approach utilizes a nonlinear MPC (NMPC) method. We use the following modified version of a discrete-time unicycle model that describes the kinematics of a nonholonomic wheeled mobile robot (WMR):
\begin{equation}\label{Eq:DiscreteModel}
\begin{split}
\mathbf{x}(k+1) &= \mathbf{x}(k) + \mathbf{f}(\mathbf{x}(k),\mathbf{u}(k))T_{s} \\
\quad \mathbf{f}(\mathbf{x}(k),\mathbf{u}(k)) &= \begin{bmatrix}
   \cos(\theta(k)) & -a\sin(\theta(k)) \\
   \sin(\theta(k)) & a\cos(\theta(k)) \\
    0 & 1
   \end{bmatrix}\begin{bmatrix}
   v \\
   \omega \\
   \end{bmatrix}
\end{split}
\end{equation}
where $k$ denotes the time step; $T_{s}$ is the sampling time; $a$ is a small positive constant; the state vector $\mathbf{x}(k) = [x(k)\quad y(k)\quad \theta(k)]^{\text{T}}$ is the robot's pose, i.e., its position $\bar{\mathbf{x}} = [x \quad y]^{\text{T}}$ and heading angle $\theta$ in the global coordinate frame at time step $k$; and the control input vector $\mathbf{u}(k) = [v(k) \quad \omega(k)]^{\text{T}}$ contains the robot's control inputs, which are its linear velocity $v$ and angular velocity $\omega$ at time step $k$. 
{If one uses the standard unicycle kinematic model of a nonholonomic WMR as in~\cite{9258372}, then the angular velocity of the robot does not show up in the barrier constraint~\cite{xiao2019control}. Thus, we use the modified kinematic model in Eq.~\eqref{Eq:DiscreteModel} so that the system has relative degree $1$.}

In an NMPC method, we first solve a nonlinear constrained optimization problem that minimizes a loss function $l(\mathbf{x},\mathbf{u})$ over a prediction horizon of $N_{p}$ time steps: 
\begin{align}\label{Eq:NMPC} 
    \mathbf{U}^{*} &= \text{argmin}_{\mathbf{u}}\sum^{N_{p}-1}_{k=0}l(\mathbf{x}(k),\mathbf{u}(k))  
    \\
    & \mathbf{x}(k+1) = \mathbf{x}(k) + \mathbf{f}(\mathbf{x}(k),\mathbf{u}(k))T_{s}
    \nonumber \\
    & \mathbf{x}_{\text{min}} \leq \mathbf{x}(k) \leq \mathbf{x}_{\text{max}}, \quad \mathbf{x}(0) = \mathbf{x}_{c}
    \nonumber \\
    & \mathbf{u}_{\text{min}} \leq \mathbf{u}(k) \leq \mathbf{u}_{\text{max}} 
    \nonumber 
\end{align}
where $\mathbf{x}_{c}$ is the robot's current odometry measurement of its pose and $\mathbf{U}^{*}\in\mathbb{R}^{2\times N_{p}}$ is a sequence of optimal control inputs for the future $N_{p}$ time steps. The bounds $\mathbf{x}_{\text{min}}$ and $\mathbf{x}_{\text{max}}$ on the state vector can be imposed to restrict the robot to move within a specific region, and the bounds $\mathbf{u}_{\text{min}}$ and $\mathbf{u}_{\text{max}}$ on the control inputs are determined by the capabilities of the robot's actuators. If the control objective is to drive the robot to a target pose $\mathbf{x}_{\text{ref}}$, then the loss function can be defined as the sum of two quadratic terms that quantify the normed distance of the robot from the target pose and the control effort:
\begin{equation}\label{Eq:SimpleNMPC}
l(\mathbf{x}(k),\mathbf{u}(k)) = ||\mathbf{x}(k) - \mathbf{x}_{\text{ref}}||^{2}_{\mathbf{Q}} + || \mathbf{u}(k)||_{\mathbf{R}}^{2},    
\end{equation}
where $\mathbf{Q}$ and $\mathbf{R}$ are square weighting matrices and $||\mathbf{x}||^{2}_{\mathbf{A}} \equiv \mathbf{x}^T\mathbf{A}\mathbf{x}$. Given the optimal control inputs $\mathbf{U}^{*}$, only the first control input $\mathbf{U}^{*}(0)$ is applied to the robot's actuators. Then, the time step is incremented from $k$ to $k+1$, and optimization problem \eqref{Eq:NMPC} is solved again. We use the NMPC formulation in Eq.~\eqref{Eq:SimpleNMPC} without stabilizing terminal costs or terminal constraints, in order to speed up the convergence rate and reduce the computation time (see~\cite{mehrez2017predictive}).

\subsection{Control Barrier Functions (CBFs)}

We define \textit{safety} as a criterion that prevents the control inputs from driving the robot into a collision with other robots or obstacles. Control barrier functions  enable safety for control synthesis by providing forward invariance property of a specified {\it safe set} based on a Lyapunov-like condition. We define a safe set $\mathcal{C}$ that represents the free space of the domain, where the robot can move without colliding with an obstacle. This set is described by the superlevel set of a continuous differentiable function $h(\bar{\mathbf{x}}(k))$, which is known as a barrier function~\cite{ames2019control,ames2014control}: 
\begin{equation}
    \mathcal{C} = \{\forall k\in\mathbb{Z}_{0},~ \bar{\mathbf{x}}(k)\in \mathbb{R}^{n}|~h(\bar{\mathbf{x}}(k)) \geq 0\}.
\end{equation}
In order to prevent collisions, the control inputs must maintain the robot's position $\bar{\mathbf{x}}$ within the safe set $\mathcal{C}$. 
In other words, the set $\mathcal{C}$ must be \textit{forward invariant} with respect to $\mathbf{f}(\bar{\mathbf{x}}(k), \mathbf{u}(k))$ defined in Eq.~\eqref{Eq:DiscreteModel}. The closed set $\mathcal{C}$ is called forward invariant if for every $\bar{\mathbf{x}}(0) \in \mathcal{C}$, $\bar{\mathbf{x}}(k) \in \mathcal{C}$ for all $k \in \mathbb{Z}_{0}$.
Then, the safety certificate at time step $k$ during robot navigation can be encoded as a Control Barrier Condition (CBC) as follows:  
\begin{equation}\label{Eq:CBC}
	 h(\bar{\mathbf{x}}(k+1)) - h(\bar{\mathbf{x}}(k)) + \gamma h(\bar{\mathbf{x}}(k))  \geq 0,
\end{equation}
where 
$\gamma$ is a small positive value less than one, i.e., $0 < \gamma \leq 1$.
In~\cite{zeng2020safety}, it was proved that adding the CBC~\eqref{Eq:CBC} to the constraints of the MPC optimization problem~\eqref{Eq:NMPC} ensures the safety of the computed optimal control inputs.

\section{Decentralized NMPC-LBF Method}\label{Section:NMPC_LBF}

\begin{figure}[t!]
\centering
     \includegraphics[width=0.5\textwidth]{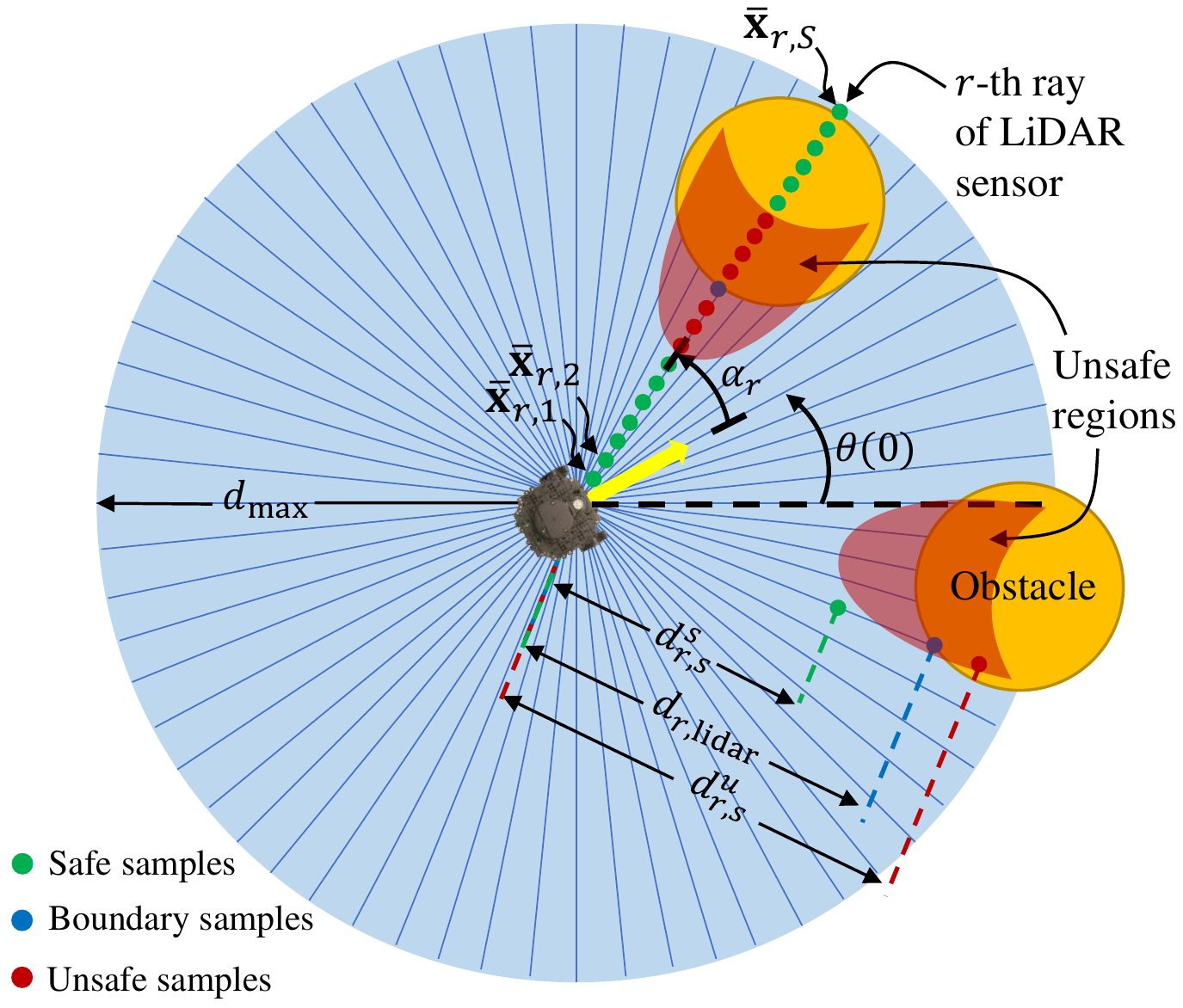}     
     \caption{LiDAR sensor rays and point sampling in safe and unsafe regions within the sensing range of the LiDAR. The unsafe samples are taken from the red shaded regions and the safe samples are in areas excluding the unsafe regions. $d^{s}_{r,s}$, $d_{r,\text{lidar}}$, $d^{u}_{r,s}$ show the distances of {the} robot to the safe, boundary, and unsafe samples on a ray of LiDAR. 
     }
    \label{Fig:SamplingSafeUnsafeRegion}
\end{figure}

In our NMPC-LBF method, the BF $h$ is learned by training a DeepNN $\hat{h}$ in real-time on each robot. 
Then, the trained DeepNN $\hat h$ is used within the NMPC constraints to provide an approximation of the BF $h$ for the finite prediction horizon of $N_p$ time steps.
\begin{figure}[t!]
\centering
     \includegraphics[width=0.5\textwidth]{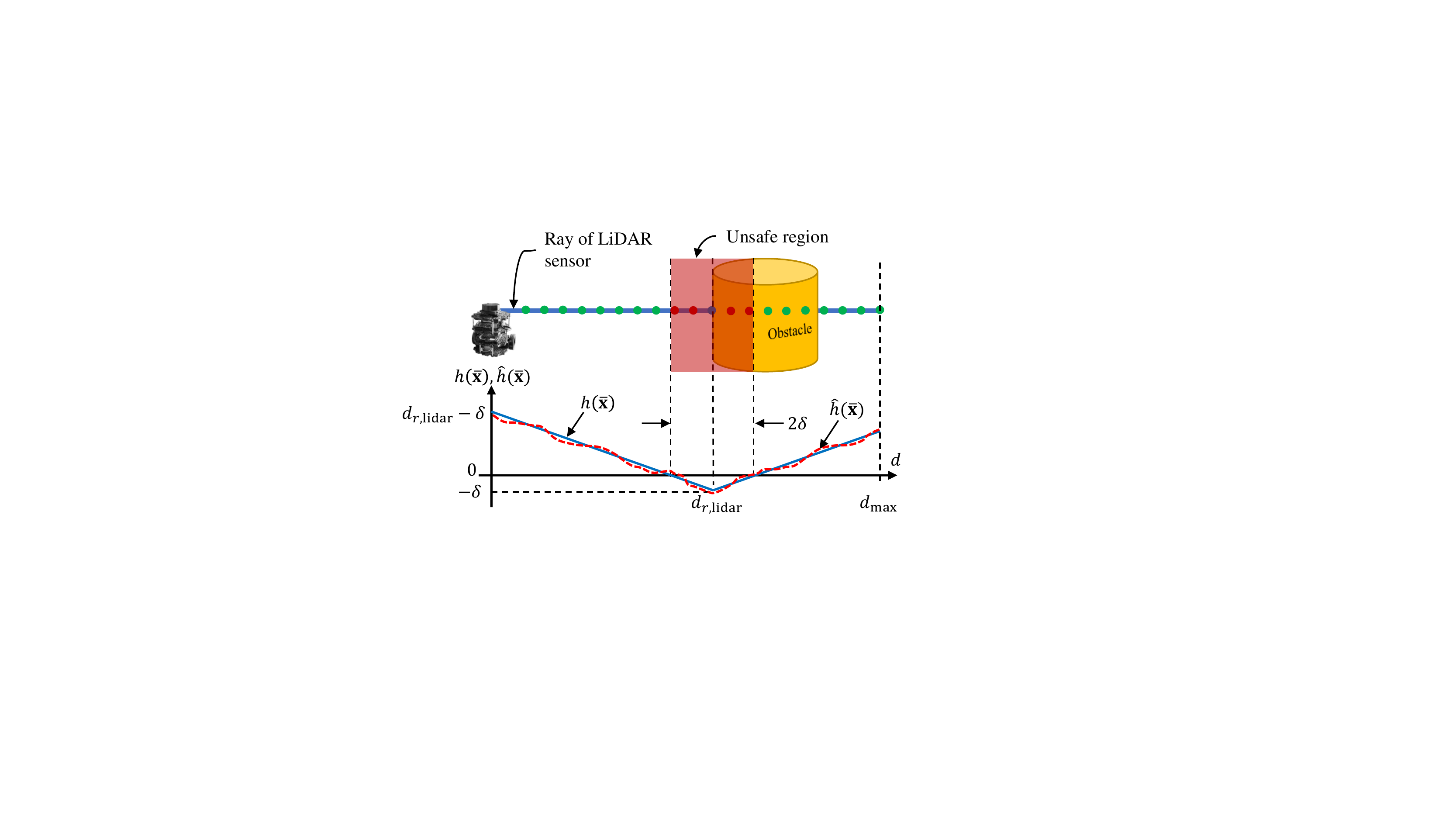}
     \caption{An illustration of ground-truth BF, $h(\bar{\mathbf{x}}$), and approximation of BF by the DeepNN, $\hat{h}(\bar{\mathbf{x}}$), on a single ray of LiDAR along with a 2D visualization of safe and unsafe samples and a sample on the boundary of {an} obstacle. We note that the sample points are defined in {the} $x-y$ plane with {a} 
     height of zero.}
    \label{Fig:BarrierFunctionRay}
\end{figure}
%

\subsection{Data Sampling and Training the DeepNN BF}

The DeepNN should be trained to learn an approximation of the BF. To train the DeepNN, we need to collect samples from safe and unsafe regions that are observed by the robot during navigation. Figure~\ref{Fig:SamplingSafeUnsafeRegion} demonstrates our method to collect samples at each time instant. 
These samples are only obtained in real-time based on the LiDAR sensor readings and the current pose of the robot via odometry readings. 

We define $R$ as the number of rays on the LiDAR sensor, $r = \{1,2,...,R\}$ as the index for the rays, $\alpha_{r}$ as the angle between the direction of the robot's heading and the $r$-th ray of the LiDAR, and $d_{r,\text{lidar}}(k)$ as the distance measured by the LiDAR in the direction of the $r$-th ray at the $k$-th time step. Then, we sample points within the sensing range of the robot along each ray of the LiDAR as illustrated in Fig.~\ref{Fig:SamplingSafeUnsafeRegion}. Let us define $d_{\text{max}}$ as the maximum distance that the robot can sense. Then, we take $S$ number of samples on each ray of the LiDAR. We define $\bar{\mathbf{x}}_{r,s}(k) \in \mathbb{R}^{2}$ as the position in the global coordinate frame of the $s$-th sample, $s \in \{1,2,...,S\}$, on the $r$-th ray at the $k$-th time step. Therefore, we have in total $N_{\text{samples}} = R \times S$ samples available in the dataset at each time step to train the DeepNN. By calculating $d_{r,s} = d_{\text{max}}s/S$ as the distance of sample position $\bar{\mathbf{x}}_{r,s}(k)$ to the origin of the robot's local frame, we can readily compute the position of the sample as:
\begin{equation}\label{Eq:SampleCalc}
\begin{split}
    \bar{\mathbf{x}}_{r,s}(k) &= \bar{\mathbf{x}}_{c} +  \mathbf{R}_{z}(\theta(k))\mathbf{d}_{r,s}(k) \\[1.5ex]
    \mathbf{R}_{z}(\theta(k)) &= \begin{bmatrix}
    \cos(\theta(k)) & -\sin(\theta(k))\\
    \sin(\theta(k)) & \cos(\theta(k))
    \end{bmatrix} \\[1.5ex]
    \mathbf{d}_{r,s}(k) &= d_{r,s}\begin{bmatrix}
    \cos(\alpha_{r}(k)) \\
    \sin(\alpha_{r}(k))
    \end{bmatrix}
\end{split}    
\end{equation}
where $\theta(k)$ is the heading angle of the robot at the $k$-th time step and $\mathbf{R}_{z}(\theta(k))$ is the standard rotation matrix that performs a rotation through angle $\theta(k)$ around about the positive $z$ axis. This rotation matrix transforms vectors described in the local frame of the robot into vectors described in the global frame $x-y$. We note that $\bar{\mathbf{x}}_{r,s}$ is located in the safe region, the boundary of an obstacle, or an unsafe region within the sensing range of the LiDAR. Thus, the input for training the DeepNN is the set of all sampled points, defined as the vector $\mathbf{X}_{s} \in\mathbb{R}^{N_{\text{samples}}\times 2}$
. We describe the ground-truth outputs as the value of the BF at the input samples:
\begin{equation}\label{Eq:BarrierFunction}
    h(\bar{\mathbf{x}}_{r,s}(k)) = |d_{r,s} - d_{r,\text{lidar}}(k)| - \delta,
\end{equation}
where $\delta > 0$ denotes half of the width of {the} unsafe region around the distance measured by the LiDAR and is set to a positive value greater than the radius of the smallest circle surrounding the robot. 

Thus, the input-output set of data for training the DeepNN can be described by $\mathcal{D} = \{\bar{\mathbf{X}}_{s}, \mathbf{H}_{s} \}$ in which each row of $\bar{\mathbf{X}}_{s} \in \mathbb{R}^{N_{\text{samples}}\times 2}$ is the position vector of the sampled point $\bar{\mathbf{x}}_{r,s}$, $s=\{1,2,...,S\}$, and each row of $\mathbf{H}_{s} \in \mathbb{R}^{N_{\text{samples}}}$ is the corresponding ground-truth output value of BF, $h(\bar{\mathbf{x}}_{r,s})$. 
%
We employ the \textit{incremental learning} method to train the DeepNN $\hat h : \mathbb{R}^2 \rightarrow \mathbb{R}$ that provides an approximation of the BF. 
Figure~\ref{Fig:BarrierFunctionRay} illustrates samples on a single ray of the robot's LiDAR and the change of the ground-truth BF and its approximation with respect to the distance to an obstacle. 

In the current implementation of our method, training data is collected and used for training the DeepNN at every iteration of the control loop.
However, other (re-)training schemes are possible -- especially, if we would like to reduce the impact of over-fitting and/or consider NN-based prediction models for the dynamic obstacles.

\subsection{Symbolic Representation of the DeepNN BF}

When an analytical representation of the CBC (\ref{Eq:CBC}) is available along with a map of the environment and motion predictions of the dynamic obstacles, then the BF conditions could be directly incorporated in the constraints of the NMPC (\ref{Eq:NMPC}).
However, in our application, we consider an unknown map. 
Therefore, we need to use a learning-based BF $\hat h$ to predict the CBC. 
As a consequence, a symbolic representation of the DeepNN $\hat h$ is necessary so that it can be used in the resulting optimization problem.

Recalling that $\bar{\mathbf{x}}(k)$, where $k\in\{0,1,...,N_{p}-1\}$, denote the future states of the robot for the NMPC finite horizon, we can obtain a symbolic expression of the approximated BF for each predicted future state, given the activation functions on each node of the DeepNN and the optimal weights after each training.
We give the symbolic expression of $\bar{\mathbf{x}}(k)$ as the input to the DeepNN and obtain a symbolic expression for the approximation of {the} BF, i.e., $\hat{h}(\bar{\mathbf{x}}(k))$, at the output of the DeepNN. 
The symbolic expression is used to generalize the representation of the approximated BF with respect to $\bar{\mathbf{x}}(k)$ that allows us to compute $\hat{h}(\bar{\mathbf{x}}(k))$ for any {possible} future 
state of the robot.
The approximated BF can be easily computed by using the Forward Propagation (FP) technique. 
The required computations in FP to obtain a symbolic expression of the approximated BF are:
%
\begin{equation}
\begin{split}
    \mathbf{A}_{l} &= \sigma(\mathbf{Z}_{l}), \quad \mathbf{Z}_{l} =  ~^{l-1}\mathbf{W}_{l}^{\text{T}}\mathbf{A}_{l-1} + \mathbf{b}_{l} \\
    \mathbf{A}_{0} &= \bar{\mathbf{x}}(k),~\mathbf{A}_{L} = \hat{h}(\bar{\mathbf{x}}(k)), \quad l = 1,2,...,L
\end{split}
\end{equation}
where $0$ is the index for the input layer, 
$l$ is the index for hidden layers, 
$L$ denotes the number of hidden layers, 
and $\sigma$ is the activation function on each node. 
The weight matrix {containing} 
the weights on the connections between two consecutive layers $l-1$ and $l$ 
is defined by $^{l-1}\mathbf{W}_{l}$. Moreover, $\mathbf{Z}_{l}$ is an affine transformation on the previous layer's output $\mathbf{A}_{l-1}$, and $\mathbf{A}_{l}$ describes the output of {layer} $l$ 
after applying the activation function on $\mathbf{Z}_{l}$. 

{\bf Remark:}
In some cases, we may want to use the continuous time formulation of BF \cite{ames2019control}, e.g., in an approximation method.
In continuous time, the equivalent CBC condition requires the derivative of the BF.
A symbolic representation of the {gradient} of the approximated BF, i.e., $\nabla \hat{h}(\bar{\mathbf{x}}(k))$, is not as straightforward as deriving a symbolic representation for $\hat{h}(\bar{\mathbf{x}}(k))$.
However, it is still possible to compute it.
To do so, we can use the Back Propagation (BP) technique to calculate a symbolic expression for the gradient 
with respect to the DeepNN's inputs, i.e., $\bar{\mathbf{x}}(k)$. 
For example, if we assume that $\sigma(x) = \tanh(x)$, then through BP, we obtain a symbolic expression for the approximated gradient of the BF with respect to the inputs by:
\begin{equation}
\begin{split}
    \frac{\partial\hat{h}(\bar{\mathbf{x}}(k))}{\partial\bar{\mathbf{x}}(k)} &= \frac{\partial\mathbf{A}_{L}}{\partial\mathbf{Z}_{L}}.\frac{\partial\mathbf{Z}_{L}}{\partial\mathbf{A}_{L-1}} ~...~\frac{\partial\mathbf{A}_{1}}{\partial\mathbf{Z}_{1}}.\frac{{\partial}\mathbf{Z}_{1}}{\partial\bar{\mathbf{x}}(k)} \\
    \frac{\partial\mathbf{A}_{l}}{\partial\mathbf{Z}_{l}} &= 1 - \tanh^{2}(\mathbf{Z}_{l}), \quad \frac{\partial\mathbf{Z}_{l}}{\partial\mathbf{A}_{l-1}} = ~^{l-1}\mathbf{W}_{l}^{\text{T}}.
\end{split}
\end{equation}
%

\subsection{Incorporating BF into NMPC}
Given the BF approximated by the DeepNN, we define sets that correspond to the safe region, its boundary, and the unsafe region of the environment, respectively:
\begin{equation}\label{Eq:SafeBoundaryUnsafeSets}
\begin{split}
    \mathcal{C} &= \{ \bar{\mathbf{x}}~|~\hat{h}(\bar{\mathbf{x}}) > \delta \}, ~~
    \partial\mathcal{C} = \{ \bar{\mathbf{x}}~|~\hat{h}(\bar{\mathbf{x}}) = \delta \}, \\
    \mathcal{U} &= \{\bar{\mathbf{x}}~|~\hat{h}(\bar{\mathbf{x}}) < \delta\}
\end{split}    
\end{equation}

\begin{figure}[t!]
\centering
     \includegraphics[width=0.43\textwidth]{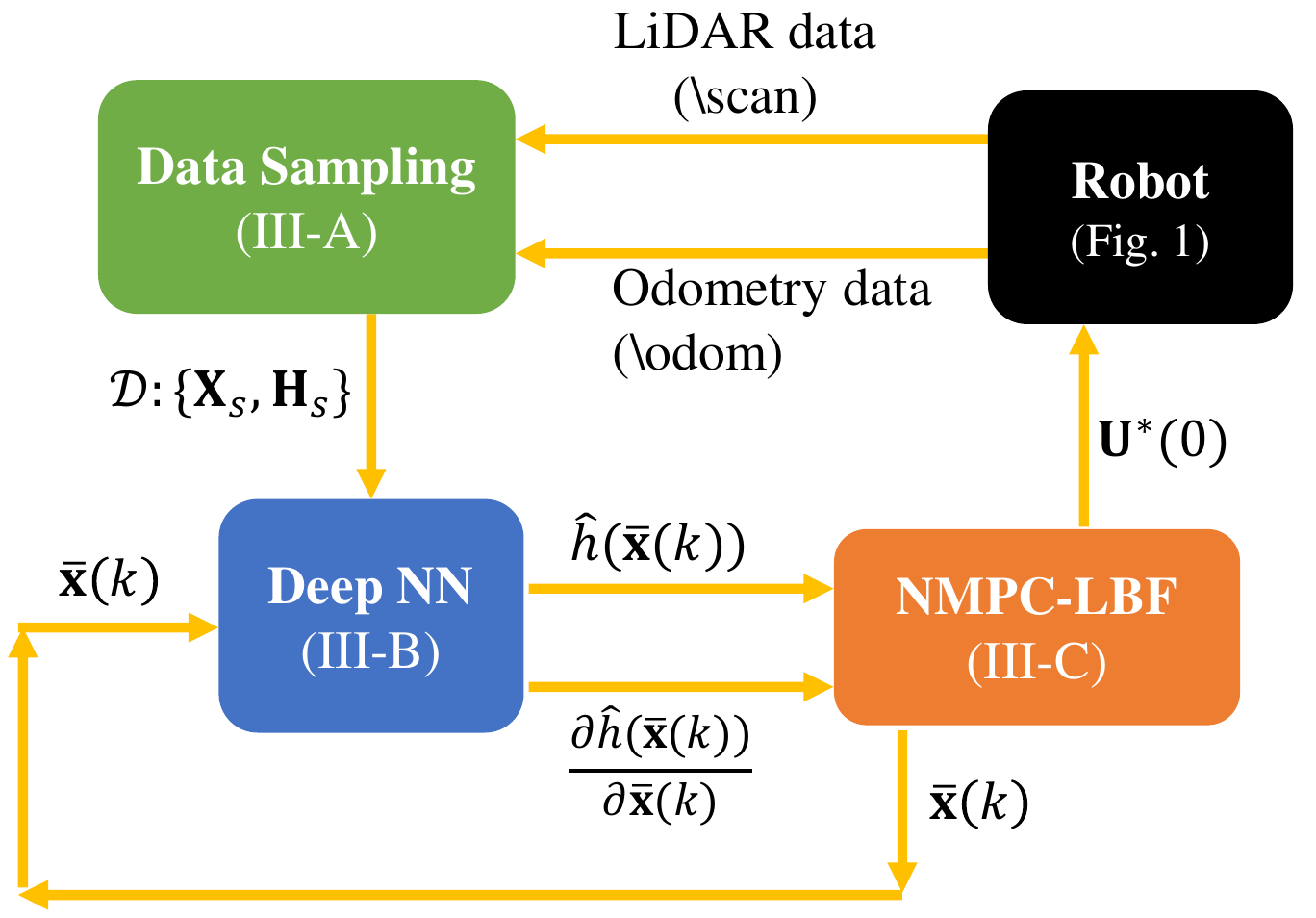}
     \caption{Block diagram of the NMPC-LBF method described in Eq.~\eqref{Eq:NMPC-LBF}. \textbackslash scan and \textbackslash odom are ROS topics through which the states of the robot and the LiDAR measurements would be available.}
    \label{Fig:BlockDiagram}
\end{figure}

For all future $N_{p}$ time steps, we calculate $\hat{h}(\bar{\mathbf{x}}(k))$  via FP computations through the DeepNN, 
which also determines whether each future state will be in the set $\mathcal{C}$, $\partial\mathcal{C}$, or $\mathcal{U}$. 
The constrained optimization problem of our NMPC-LBF method is formulated as: 
\begin{align}\label{Eq:NMPC-LBF} 
    \mathbf{U}^{*}, \mathbf{X}^{*} &= \text{argmin}_{\mathbf{x},\mathbf{u}}\sum^{N_{P}-1}_{k=0}l(\mathbf{x}(k),\mathbf{u}(k))  
     \\
    & \mathbf{x}(k+1) = \mathbf{x}(k) + \mathbf{f}(\mathbf{x}(k),\mathbf{u}(k)){T_s}
    \nonumber \\
    & \mathbf{x}_{\text{min}} \leq \mathbf{x}(k) \leq \mathbf{x}_{\text{max}}, \quad \mathbf{x}(0) = \mathbf{x}_{c} 
    \nonumber \\
    & \mathbf{u}_{\text{min}} \leq \mathbf{u}(k) \leq \mathbf{u}_{\text{max}} 
    \nonumber \\
    &  \hat h(\bar{\mathbf{x}}(k+1)) - \hat h(\bar{\mathbf{x}}(k)) + \gamma \hat h(\bar{\mathbf{x}}(k))  \geq 0 ~(\text{CBC})
    \nonumber 
\end{align}
where $\mathbf{X}^{*}$ is a sequence of optimal states. 
In order to reduce the computational complexity in this optimization problem, we use {the} multiple shooting method and lift up the problem by considering both $\mathbf{x}$ and $\mathbf{u}$ as decision variables in the minimization. 
We implement and solve this constrained nonlinear optimization problem via CasADi, {which} 
is a powerful framework specialized for solving NMPC problems by providing a symbolic expression of the problem. 
In the decentralized framework of our method, each robot solves the optimization problem described in Eq.~\eqref{Eq:NMPC-LBF} independently in which the BF is being learned by the DeepNN online in the loop at each time step. Figure~\ref{Fig:BlockDiagram} shows a block diagram of the NMPC-LBF method.  
%

\subsection{Implementation and Parameter Tuning}

We describe the implementation of our method using the pseudo codes in Algorithms 1 and 2. Algorithm 1 presents the NMPC-LBF method step by step. After initialization of the problem and formulating the optimization problem (line 1), the main loop (line $2$ to $9$) is executed online to collect the samples and train the DeepNN as described in Algorithm 2 given the 
robot's {current} pose from odometry and LiDAR measurements. We obtain optimal control inputs after solving Eq.~\eqref{Eq:NMPC-LBF} and apply {them} to the robot. Then, we shift the prediction horizon and initialize the states and control inputs in the optimization problem with a warm start. This loop is executed up to the point that the distance of the robot to its goal becomes less than $e_{\text{ref}}=0.1$. 

There are several parameters in the NMPC-LBF method that should be tuned carefully to achieve the expected performance of the robot to safely navigate in the environment. To implement the NMPC-LBF method, we recommend the following values for parameters to achieve the desired performance: 
prediction horizon 
$N_{p}=10\sim 20$, sampling time 
$T_{s}=0.01\sim 0.05$ {s}, weight matrices 
$\mathbf{Q}=\text{diag}(5,5,0.05)$ and $\mathbf{R}=\text{diag}(2,0.5)$, learning rate 
$l_{r}=0.01$, number of samples on each ray of LiDAR $S = 50$, half width of unsafe region $\delta = 0.2$ m, number of epochs $n_{\text{epochs}}=20$ in the training of DeepNN, and $\beta = 0.1\sim 0.2$ in the CBC.
We use a fully-connected DeepNN that is implemented in TensorFlow~\cite{abadi2016tensorflow} and Keras~\cite{chollet2015keras} with $2$ neurons at the input,
and one output.
The network architecture for the hidden layers is $n_{l}=\{32, 32, 16, 16, 8\}$, and we use $\tanh$ as the activation function for all nodes. 

%
\begin{algorithm}[t!]
  \caption{NMPC-LBF Method}\label{algorithm1}
   \textbf{Input:} $\mathbf{x}(0)$, $\mathbf{x}_{\text{ref}}$, $\mathbf{Q}$, $\mathbf{R}$, $\mathbf{x}_{\text{min}}$, $\mathbf{x}_{\text{max}}$, $\mathbf{u}_{\text{min}}$, $\mathbf{u}_{\text{max}}$, $T_{s}$, $N_{p}$
   \\ [1ex]
   \textbf{Output:} $\mathbf{U}^{*}(0)$ \\[-1.75ex]
\begin{algorithmic}[1]
\STATE Initialize ROS node, odometry and LiDAR scan subscribers, and velocity command publishers{;} 
symbolically formulate Eq.~\eqref{Eq:NMPC-LBF} using CasADi 
\WHILE{$||\mathbf{x}(k)-\mathbf{x}_{\text{ref}}(k)|| > e_{\text{ref}}$}
      \STATE Obtain {robot's} 
      pose and LiDAR measurements 
      \STATE Algorithm 2: Data Sampling and DeepNN Training
      \STATE Solve optimization problem in Eq.~\eqref{Eq:NMPC-LBF} to obtain $\mathbf{U}^{*}$
      \STATE Publish $\mathbf{U}^{*}(0)$ to ROS topics of {robot's} 
      velocity commands
      \STATE Update initial guess: $\mathbf{U}(0) \leftarrow \mathbf{U}^{*}(0)$, $\mathbf{X}(0) \leftarrow \mathbf{x}_{c}$
      \STATE Increment time step: $k \leftarrow k+1$ 
\ENDWHILE
\end{algorithmic}
\end{algorithm}
%
\begin{algorithm}[t!]
  \caption{Data Sampling and DeepNN Training}\label{algorithm2}
   \textbf{Input:} $\mathbf{x}_{c}$, $\mathbf{x}(k)$, $d_{r,\text{lidar}}$, $d_{\text{max}}$, $R$, $S$, $l_{r}$, $n_{\text{epochs}}$\\ [1ex]
   \textbf{Output:} $\hat{h}(\mathbf{x}(k)), \partial\hat{h}(\mathbf{x}(k))/\partial\mathbf{x}(k)$\\[-1.75ex]
\begin{algorithmic}[1]
\STATE $\bar{\mathbf{X}}_{s} = [~]$, $r = 0$, $s = 0$
\FOR{$r < R$}
      \STATE Compute $d_{r,s}$, $\alpha_{r}$ given $r$ and $s$
      \FOR{$s < S$}
      \STATE Compute $\mathbf{d}_{r,s}$ in Eq.~\eqref{Eq:SampleCalc}
      \STATE Obtain sample positions {$\bar{\mathbf{x}}_{r,s}$} 
      in Eq.~\eqref{Eq:SampleCalc}
      \STATE Collect data samples $\bar{\mathbf{X}}_{s} \leftarrow \bar{\mathbf{x}}_{r,s}$ 
\ENDFOR
\ENDFOR
\STATE Calculate and return $\hat{h}(\bar{\mathbf{x}}(k))$ via FP
\end{algorithmic}
\end{algorithm}
%

\section{Simulation and Experimental Results}\label{Section:Simulation_Experiments}

To evaluate the effectiveness of our method, we implemented it on simulated TurtleBot3 (TB3) Burger robots~\cite{turtlebot_website} in Gazebo and on an actual TB3 Burger robot. 

For the simulation analysis, we simulated different scenarios for navigation of single and multiple robots in unknown environments. Due to space limitations, we only present results for two of these scenarios in this paper. Videos of the experiments with the real robot and all simulations, including the simulations not discussed here, are available at~\cite{VideosSimExp}. 

\begin{figure}[t!]
\centering
     \includegraphics[width=0.475\textwidth]{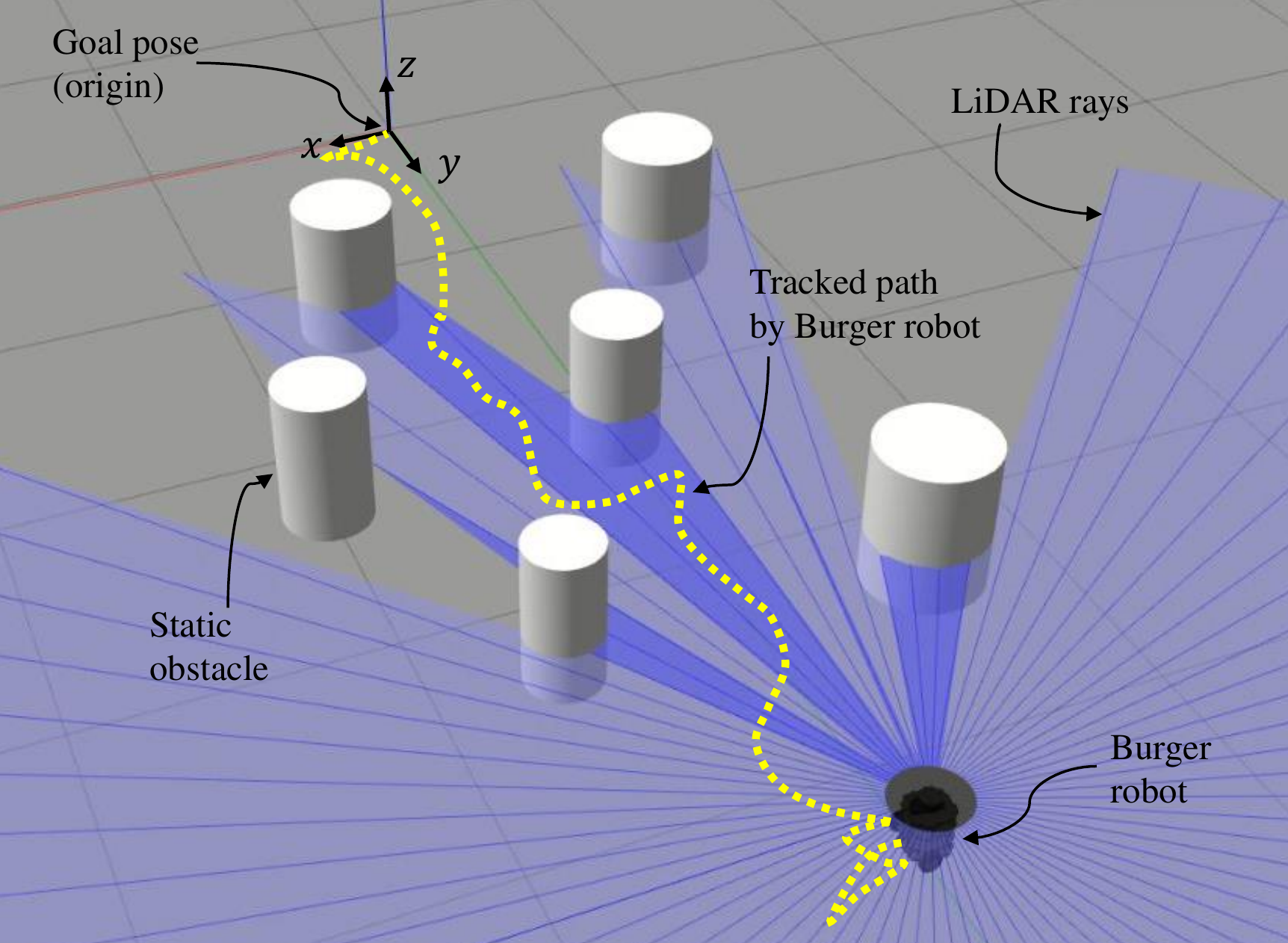}
     \caption{A snapshot of the simulation of scenario $1$ in Gazebo. The robot should stabilize to the goal position at the origin of the global frame.}
    \label{Fig:Scenario1--Snapshot}
\end{figure}

\begin{figure}[t!]
\centering
     \includegraphics[width=0.5\textwidth]{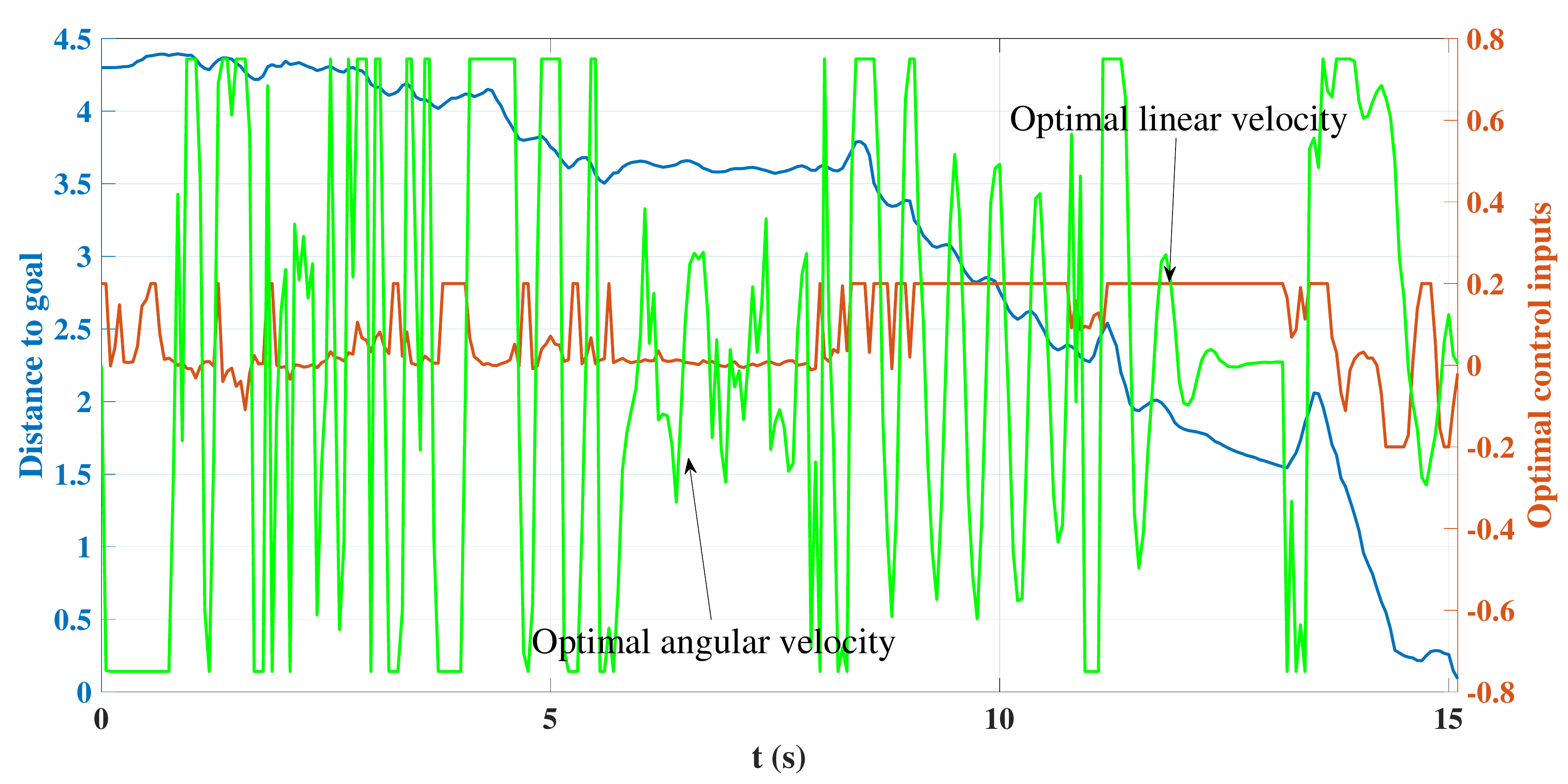}\vspace*{-5mm}
     \caption{
     Distance to goal position and optimal control inputs over time of the robot in scenario 1.} 
    \label{Fig:Scenario1--graphs}
\end{figure}

In scenario $1$, a single robot should stabilize to a goal position in an environment with six unknown static obstacles. 
Figure~\ref{Fig:Scenario1--Snapshot} shows a snapshot of the initial configuration of the robot and its trajectory during a simulation of this scenario.
The distance of the robot to the goal position and the optimal control inputs over time are presented in Figure~\ref{Fig:Scenario1--graphs}. In our previous work~\cite{9258372}, this control objective was achieved in the same environment 
using gradient-based feedback controllers 
that require 
the robot 
to have a priori knowledge about the positions and geometry of the obstacles. 
In contrast, 
the NMPC-LBF method can perform online learning in an unknown environment with dynamic obstacles and multiple robots. 
This property of the NMPC-LBF method allows its real-time deployment 
on actual robots.

\begin{figure}[t!]
\centering
     \includegraphics[width=0.475\textwidth]{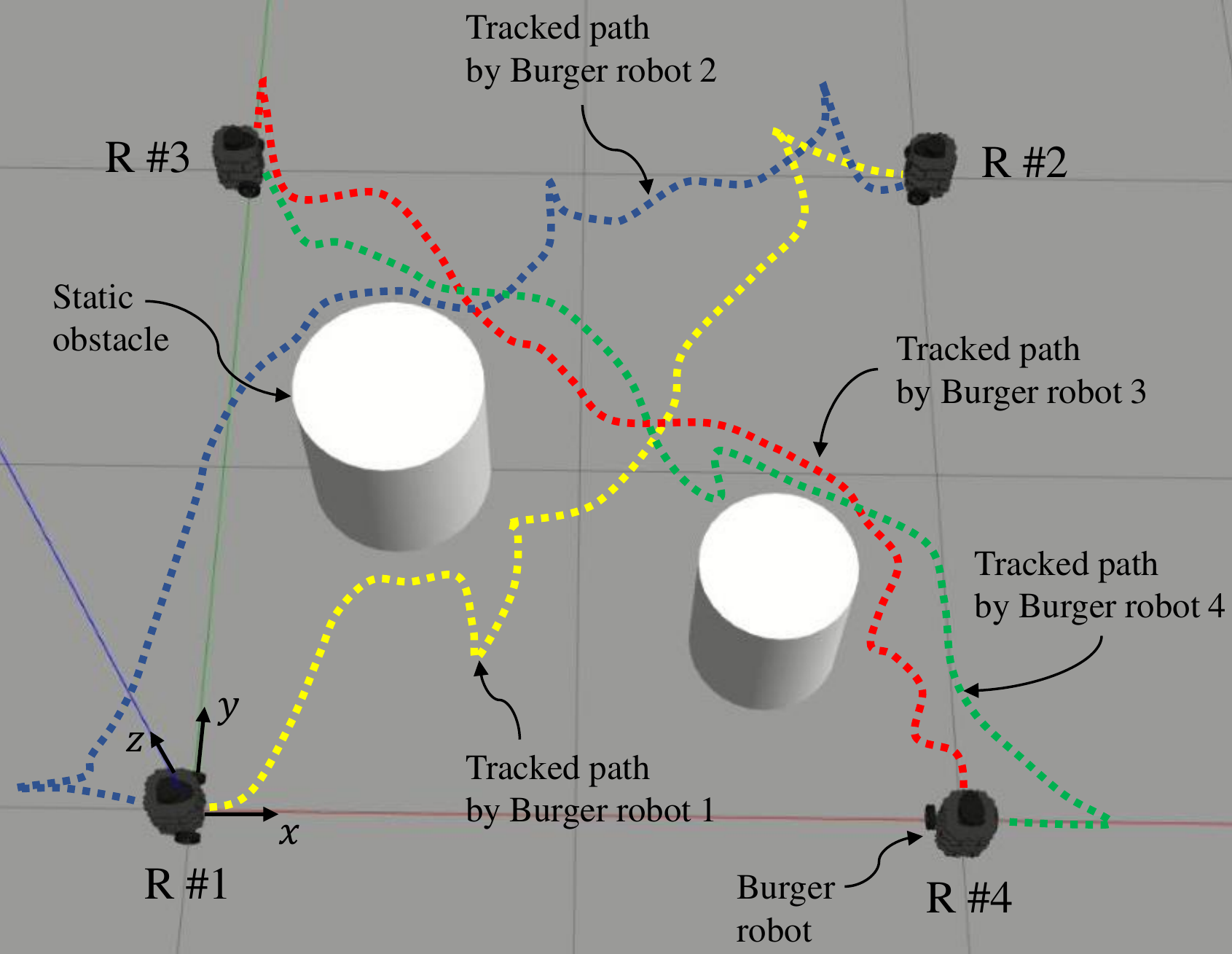}
     \caption{A snapshot of the simulation of 
     scenario $2$ in Gazebo. All robots should stabilize to their goal positions while avoiding collisions with other robots and static obstacles.}
    \label{Fig:Scenario3--Snapshot}
\end{figure}

\begin{figure}[t!]
\centering
     \includegraphics[width=0.5\textwidth]{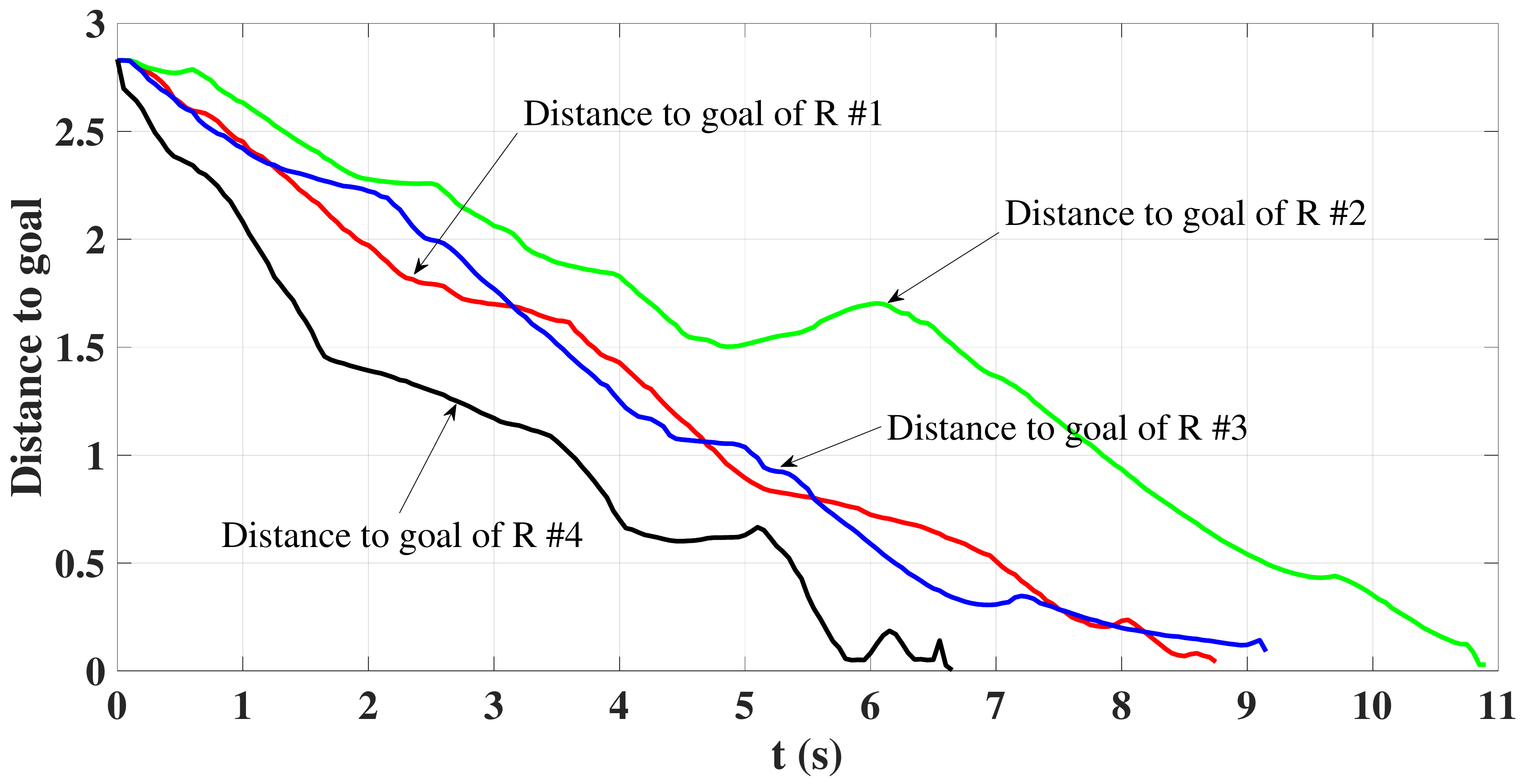}\vspace*{-5mm}
     \caption{Distances of the robots to their goal positions over time in scenario 2. 
}
    \label{Fig:Scenario3--graphs}
\end{figure}

In scenario $2$, four robots should stabilize to their goal positions while avoiding collisions with one another and two unknown static obstacles in the environment. 
Figure~\ref{Fig:Scenario3--Snapshot} shows a snapshot of the initial configuration of the robots (R $\#i, \quad i=\{1,2,3,4\}$) and their trajectories during a 
simulation of this scenario. 
Plots of the 
distances of the robots to their goal positions over time 
are shown in Fig.~\ref{Fig:Scenario3--graphs}. 

In experimental tests (see the video~\cite{VideosSimExp}), we implemented the NMPC-LBF method on a single robot in three different scenarios in which the robot should stabilize to its goal position in an environment with one or two unknown 
obstacle(s). Two of the scenarios have static obstacle(s), and the third has a dynamic obstacle.
{To implement the controller, 
we used a computer with Linux Ubuntu 16.0, Intel Core i5 processor, 8GB memory, and ROS Kinetic. The average computation time for solving the NMPC-LBF is $0.25$ s and the update rate of the DeepNN weights is $0.15$ s.} 
{The NMPC-LBF code in Python is available at~\cite{nmpc-lbf-python}}.
%
%
 
%
%

\section{Analysis and Discussion}

\subsection{Stability and Feasibility}

The stability of the proposed control approach can be established using proofs similar to those in our previous work~\cite{lafmejani2021nonlinear} if we can show that the associated optimization problem in Eq.~\eqref{Eq:NMPC-LBF}, with CBCs as constraints, has recursive feasibility~\cite{lars2011nonlinear}. Let us denote $\mathbb{X}$ and $\mathbb{U}$ as the feasible sets of states and control inputs for the robot. We use the simplified notation $\mathbb{Z} = \mathbb{X} \times \mathbb{U}$. Suppose that $\bar{\mathbf{x}}(0)$ is known and is in a feasible state, i.e., $\bar{\mathbf{x}}(0) \in \mathbb{X}$. Then, the optimization problem in Eq.~\eqref{Eq:NMPC-LBF} has feasible solutions if and only if for all time steps $0, ...,N_{p}-1$: \textbf{(a)} the optimal control solutions and corresponding predicted states are a subset of the feasible set, i.e., $\mathbf{X}^{*}\times\mathbf{U}^{*}\subseteq \mathbb{Z}$, and \textbf{(b)} the CBCs in Eq.~\eqref{Eq:CBC} are satisfied.

To prove statement $\textbf{(a)}$, we consider the kinematic model of each robot in Eq.~\eqref{Eq:DiscreteModel}. If the initial state and initial control input are in the feasible set, i.e.,  
$\bar{\mathbf{x}}(0)\times\mathbf{u}(0) \in\mathbb{Z}$, then the recursive feasibility of the optimization problem in Eq.~\eqref{Eq:NMPC-LBF} trivially holds. 
To establish statement $\textbf{(b)}$, 
we would need two conditions:
(i) $\hat h$ is a ``good" approximation of $h$, and
(ii) the sampling rate $T_s$ is sufficient.
Even though condition (ii) can be guaranteed given the dynamics of the robots and the obstacles,  condition (i) remains 
challenging 
to demonstrate, 
in general. 
Using NN verification methods, e.g., \cite{DuttaEtAl2019hscc,TranEtAl2020cav}, we could establish that the approximation error is bounded.
If the error is bounded, then we could guarantee that any control input that satisfies the CBC in the optimization problem in Eq. (\ref{Eq:NMPC-LBF}) will also satisfy the CBC \eqref{Eq:CBC}.
This will be the focus of our future work.

\subsection{Limitations of Proposed NMPC-LBF Method}
As a limitation of our decentralized method, a powerful computational resource on the robot is required to train the DeepNN and solve the optimization problem of NMPC-LBF in real-time. In turn, this allows us to increase the resolution of sampling points on each ray of the LiDAR to collect more data, which could preempt the over-fitting problem in training of the DeepNN. Another limitation 
is the challenging task of tuning the parameters of the NMPC-LBF method, which significantly affects the safety and stability of the robot during navigation. Although inter-robot collision avoidance can be ensured in our method, there are some special cases in which collision avoidance between the robot and dynamic obstacles is not guaranteed. If the velocity vector of a dynamic obstacle aligns with the velocity vector of the robot and has a greater magnitude, then they might collide since the robot's control inputs are constrained and there are no known constraints on the obstacle's dynamics. 

Due to the generalization error in learning of the BF by the DeepNN, there always exists a difference between the learned BF and its 
ground-truth values. 
Given the ground-truth values of the BF, computing the approximation error for the training sample positions would be possible, whereas we are not able to compute it 
for test sample positions that are not in our dataset. 
This issue could be addressed by using an NN verification method, e.g., \cite{DuttaEtAl2019hscc,TranEtAl2020cav}, to bound the approximation error.
Lastly, the NMPC-LBF method does not necessarily generate smooth trajectories while approaching the obstacles. 
This issue could be solved by using a stochastic NMPC method as described in~\cite{park2016graceful} to generate graceful motions of the robot during navigation.

\section{Conclusions and Future Work}\label{Sec:Conclusion}

In this paper, we presented a decentralized control approach based on an NMPC method leveraged by a DeepNN to learn BF to ensure safety for navigation of mobile robots in unknown environments in the presence of other robots and obstacles. The proposed method does not require inter-robot communication and it can be scaled up to be implemented on any number of robots. Future work includes modifying the 
method for learning the BF over the prediction horizon, and therefore estimating the unsafe regions at future time steps, by using 
a history of the robots' LiDAR readings as inputs to the DeepNN. 
Another direction for future work is to redesign the optimization problem in order to encourage smoothness in the robots' navigation. 

\bibliographystyle{elsarticle-num}
\bibliography{iros2022.bib}

\end{document}